%% file: arxiv.tex
\documentclass{article} 
\usepackage{iclr2027_conference,times}

\iclrfinalcopy
\input{math_commands.tex}

\usepackage{hyperref}
\usepackage{url}
\usepackage{graphicx}
\usepackage{booktabs}
\usepackage{multirow}
\usepackage{multicol}
\usepackage{dblfloatfix}
\usepackage{enumitem}
\usepackage{pifont}
\usepackage{caption}
\usepackage{float}
\usepackage{xcolor}
\usepackage{fontawesome5}
\usepackage{hyperref}
\definecolor{titleblue}{RGB}{28,42,168}
\hypersetup{colorlinks=true,urlcolor=titleblue}

\usepackage{pifont}

\definecolor{checkgreen}{RGB}{46,139,87}
\definecolor{crossred}{RGB}{200,65,65}

\newcommand{\cmark}{\textcolor{checkgreen}{\ding{51}}}
\newcommand{\xmark}{\textcolor{crossred}{\ding{55}}}

\title{WorldMirror: Agentic Real-to-Sim Generation of Interactive Worlds from In-the-Wild Robot Demonstrations}

\author{%
\parbox[t]{\dimexpr\textwidth-12pt\relax}{%
\mdseries
Tianling Xu$^{1,*}$, Weijie Wan$^{1,*}$, Honghao Su$^{1,*}$, Kailun Su$^{1}$, Junyuan Tang$^{1}$, Huichi Zhou$^{4}$\\[0.25em]
Haozheng Zhang$^{4}$, Qi Xiong$^{1,2}$, Ping Luo$^{3}$, Wenbo Ding$^{1,2,\dagger}$, Tianxing Chen$^{1,3,\dagger}$, Yue Chen$^{1,5,\dagger}$\\[0.35em]
\mbox{$^{1}$Xspark AI}\quad
\mbox{$^{2}$Tsinghua University}\quad
\mbox{$^{3}$The University of Hong Kong}\\
\mbox{$^{4}$University College London}\quad
\mbox{$^{5}$Peking University}\\[0.45em]
Project page: \url{https://worldmirror-page.github.io/}%
}}

\begin{document}

\maketitle

\begingroup
\renewcommand{\thefootnote}{}%
\footnotetext{*Equal contribution. \quad $\dagger$Corresponding author.}
\addtocounter{footnote}{-1}%
\endgroup


\input{sections/0_abstract}
\input{figure/teaser}
\input{sections/1_introduction}

\input{sections/2_related-work}
\input{sections/3_method}
\input{sections/4_experiment}
\input{sections/5_conclusion}

\newpage
\bibliography{iclr2027_conference}
\bibliographystyle{iclr2027_conference}


\end{document}

%% file: math_commands.tex
\usepackage{amsmath,amsfonts,bm}

\def\eqref#1{equation~\ref{#1}}

\def\1{\bm{1}}

\DeclareMathAlphabet{\mathsfit}{\encodingdefault}{\sfdefault}{m}{sl}
\SetMathAlphabet{\mathsfit}{bold}{\encodingdefault}{\sfdefault}{bx}{n}



%% file: sections/0_abstract.tex
\begin{abstract}
 Existing approaches for robot learning in simulation often require substantial manual effort to construct scenes and specify tasks, limiting training scalability and data diversity. Meanwhile, in-the-wild robot demonstrations naturally provide diverse and realistic observations of real-world interactions. Leveraging these demonstrations for simulation requires both high-fidelity visual reconstruction and physically consistent interactions. General-purpose agents offer flexible adaptation to diverse scenarios and strong reasoning about 3D spatial relationships, while specialized generative models provide high-fidelity reconstruction of scene appearance and geometry. We introduce \textbf{WorldMirror}, an agentic real-to-sim framework that constructs interactive digital twins from in-the-wild robot demonstrations. Our central idea is to formulate real-to-sim construction as the iterative refinement of an executable world program. A coding agent decomposes the observed scene, plans reconstruction steps, and orchestrates specialized image and 3D generation tools, refining the program through verification feedback. A camera-pose-anchored scene representation organizes the world into static visual context, support geometry, and editable interactive assets while preserving spatial relationships among the scene, objects, and robot. The agent uses this representation to diagnose and correct reconstruction errors, enabling physically consistent interaction and trajectory replay.
 Experiments on DROID demonstrate that WorldMirror can reconstruct visually faithful and geometrically consistent interactive worlds while preserving the physical validity of recorded robot trajectories.
 

\end{abstract}

%% file: figure/teaser.tex

\begin{figure}
    \centering
    \includegraphics[width=\linewidth]{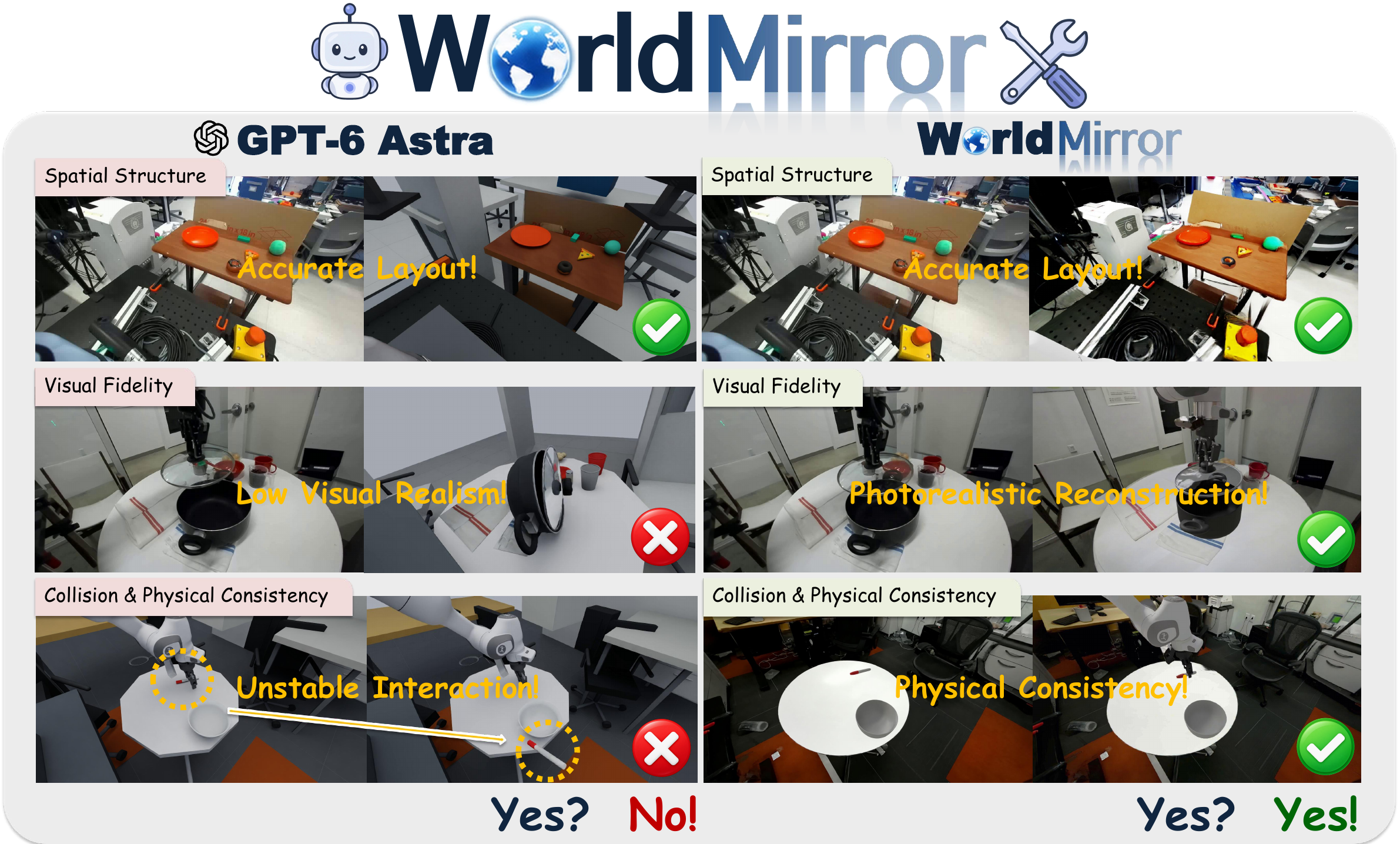}
\caption{
WorldMirror constructs simulation-ready digital twins with high visual and interaction fidelity.
GPT-6 Astra is competitive in spatial structure and layout,
while WorldMirror achieves better visual fidelity, collision geometry, and physical consistency through agentic reconstruction.
}

    \label{fig:teaser}
\end{figure}

%% file: sections/1_introduction.tex
\section{Introduction}
\label{sec:introduction}

Scaling generalist robot policies requires diverse training data across a wide range of real-world scenarios~\citep{intelligence2025pi05,black2024pi0,wang2026openwam}.
Yet collecting real-world demonstrations and testing policies across environments remain costly and labor-intensive.
Simulation offers a complementary route to scalable data generation and controlled, repeatable evaluation~\citep{li2024simpler,jain2025polaris}.
Realizing this potential requires interactive environments that reproduce the appearance and physical interactions of real-world deployment settings.

For simulation to serve these purposes, scene geometry, object placement, camera viewpoints, and robot configuration must match the real setup.
Even small errors in scale or contact geometry can alter task outcomes.
Recent real-to-sim systems have advanced scene reconstruction and its use in robot learning and evaluation~\citep{zhao2025rola,ranawaka2026simfoundry,jain2025polaris,zhang2026robosnap,dai2024acdc,torne2024rialto}.
However, real-to-sim construction involves coupled visual, geometric, and physical constraints that an feed-forward reconstruction may not fully satisfy.
Meanwhile, general-purpose agents alone may not produce the fine-grained geometry and photorealistic appearance required for faithful scene reconstruction.
Different reconstruction errors require different forms of correction.
A misplaced object may require layout adjustment, whereas a missing object may require revisiting scene decomposition and asset reconstruction.
Unsuitable collision geometry may instead require revising an asset's physical representation.
Addressing these errors calls for an adaptive and iterative correction process that selects and revisits reconstruction steps based on the current world state and verification feedback, rather than following a fixed sequence of tool calls. 
These challenges motivate \emph{agentic world modeling}, in which a coding agent coordinates specialized reconstruction tools and uses verification feedback to refine an executable world.

We introduce \textbf{WorldMirror}, an agentic real-to-sim framework that converts robot videos and their associated trajectories into interactive digital twins.
At its core, a coding agent decomposes the reconstruction objective into subtasks and coordinates specialized reconstruction tools to construct an executable world program.
Using the current world state and verification feedback, the agent determines which components to revise and which tools to invoke next, updating generated assets and assembly code accordingly.
A camera-pose-anchored scene representation organizes the reconstructed world into static visual context, support geometry, and editable interactive assets.
It places the scene, objects, and robot in a shared coordinate system, providing a consistent geometric reference for replaying recorded robot trajectories without modification.
Table~\ref{tab:system_comparison} compares WorldMirror with representative real-to-sim and agentic world construction systems. We validate WorldMirror on DROID~\citep{droid2024}
against existing baselines. Our contributions are:
\begin{itemize}[leftmargin=*]
\item We propose \textbf{WorldMirror}, an agentic real-to-sim framework constructing interactive digital twins via adaptive tool orchestration and verification-guided refinement of executable world programs.
\item We introduce a camera-pose-anchored scene representation that organizes reconstructed worlds into static visual context, support geometry, and editable interactive assets, preserving spatial relationships across the scene, objects, and robot while supporting replay of robot trajectories.
\item We evaluate WorldMirror on DROID robot manipulation demonstrations
across visual quality, layout fidelity, and motion and interaction fidelity,
demonstrating competitive reconstruction quality and improved replay fidelity
on the evaluated demonstrations.
\end{itemize}

%% file: sections/2_related-work.tex
\section{Related Work}
\label{sec:related_work}

\input{table/baselines-review}
\paragraph{Real-to-sim world construction.}
Constructing interactive digital twins requires recovering both the appearance of a real scene and the physical structure needed to support robot interaction.
Existing systems approach this problem through different reconstruction strategies: RialTo builds scanned digital twins for policy improvement~\citep{torne2024rialto}, while RoLA constructs interactive environments from individual images~\citep{zhao2025rola}. PolaRiS further develops scalable real-to-sim reconstruction for generalist robot policy evaluation, using neural reconstruction to turn short video scans of real-world scenes into interactive simulation environments~\citep{jain2025polaris}.
Several methods further integrate reconstruction with explicit alignment and physical refinement.
Other approaches explicitly incorporate geometric alignment and physical refinement. SimFoundry combines video-based reconstruction, pose alignment, and physics annotation, and generates digital cousins for robot evaluation and training~\citep{ranawaka2026simfoundry}.
RoboSnap separates interactive geometry from Gaussian visual context and refines scene configurations through geometric and physics optimization~\citep{zhang2026robosnap}, while TwinAligner jointly addresses visual alignment and rigid-body system identification~\citep{fan2025twinaligner}.
These approaches establish visual reconstruction, physical geometry,
and robot-world alignment as important components of interactive
real-to-sim systems.
Building on these advances, WorldMirror organizes reconstruction
and alignment within an executable world program, where specialized
agents coordinate component-level updates guided by verification feedback.

\paragraph{Agentic 3D Modeling.}
Recent advances in multimodal agents have enabled programmatic
construction and manipulation of 3D content.
GPT-6 Astra~\citep{openai2026gpt6astra} demonstrates the use of
general-purpose agents for constructing complex 3D worlds,
while Articraft~\citep{zhou2026articraft} and
Procedura~\citep{lin2026procedura} explore agent-driven asset
generation and procedural modeling.

Agentic approaches have also incorporated specialized generation
tools and verification mechanisms for scene construction.
SAGE~\citep{xia2026sage} coordinates scene and object generators
with critics to iteratively construct simulation-ready environments.
LiteReality~\citep{huang2026litereality} reconstructs RGB-D scans into graphics-ready scenes with structured layouts, retrieved object
assets, realistic materials, and basic physical properties.
These works illustrate complementary approaches to combining
structured scene representations, specialized reconstruction
tools, and agentic generation for scalable and simulation-ready
3D world construction.

More closely related to our setting,
Agentic Real2Sim~\citep{chen2026agentic}
combines agentic reconstruction, scene calibration, and
simulator-guided refinement to reconstruct recorded robot interactions.
WorldMirror similarly addresses agentic reconstruction of
interactive environments, with an emphasis on hierarchical
construction over a shared, editable world state.
Its router coordinates specialized agents across subtasks,
while local verification and refinement support iterative updates
to reconstructed components.
The resulting world is assembled under a camera-anchored coordinate
system to support replay of recorded robot trajectories.

\paragraph{Simulation for Robot Learning}
Simulation environments provide an important foundation for scalable robot learning. RLBench offers a diverse suite of language-conditioned manipulation tasks with reusable simulation environments and demonstrations~\citep{james2020rlbench}, while robosuite provides a modular framework for constructing and evaluating robot manipulation tasks~\citep{zhu2020robosuite}. ManiSkill further scales simulation-based learning with diverse manipulation tasks, object assets, and large-scale demonstrations~\citep{gu2023maniskill2}. SIMPLER demonstrates the use of simulation as a scalable and reproducible complement to real-world evaluation, providing simulated environments for common real robot setups~\citep{li2024simpler}. MimicGen adapts a small set of human demonstrations to diverse simulated scenes and object configurations~\citep{mandlekar2023mimicgen}, while RoboTwin provides large-scale simulation environments and task variations for robot manipulation learning~\citep{chen2025robotwin}. These approaches demonstrate the potential of simulation for scalable robot learning by providing increasingly diverse simulation environments and large-scale demonstrations. WorldMirror complements these efforts by combining agentic real-to-sim reconstruction with in-the-wild robot demonstrations to construct executable interactive worlds.


%% file: table/baselines-review.tex
\begin{table}[t]
\centering
\fontsize{7.5pt}{9.5pt}\selectfont
\captionsetup{skip=6pt}
\caption{
Capability comparison of representative real-to-sim and agentic
world construction systems, covering scene reconstruction,
robot--world alignment, and agentic construction.
}
\label{tab:system_comparison}

\setlength{\tabcolsep}{2.2pt}
\renewcommand{\arraystretch}{1.35}

\begin{tabular}{@{}lccccccc@{}}
\toprule
\textbf{Capability}
& \textbf{RoLA}
& \textbf{RoboSnap}
& \textbf{SimFoundry}
& \textbf{SAGE}
& \textbf{Astra}
& \textbf{Agentic Real2Sim}
& \textbf{WorldMirror} \\
\midrule

\multicolumn{8}{l}{\textit{Scene Reconstruction}} \\
\midrule

Scene Appearance
& \cmark & \cmark & \cmark
& \cmark & \xmark & \cmark & \cmark \\

Interactive Assets
& \cmark & \cmark & \cmark
& \cmark & \cmark & \cmark & \cmark \\

\midrule
\multicolumn{8}{l}{\textit{Robot--World Alignment}} \\
\midrule

Camera Calibration
& \xmark & \cmark & \cmark
& \xmark & \cmark & \cmark & \cmark \\

Camera-Pose-Anchored
World Representation
& \xmark & \xmark & \xmark
& \xmark & \xmark & \xmark & \cmark \\

Demonstration Replay
& \xmark & \cmark & \xmark
& \xmark & \cmark & \cmark & \cmark \\

\midrule
\multicolumn{8}{l}{\textit{Agentic Construction}} \\
\midrule

Tool Orchestration
& \xmark & \xmark & \xmark
& \cmark & \cmark & \cmark & \cmark \\

Agent-based Verification
& \xmark & \xmark & \xmark
& \cmark & \cmark & \cmark & \cmark \\

Feedback-guided Refinement
& \xmark & \cmark & \cmark
& \cmark & \cmark & \cmark & \cmark \\

\bottomrule
\end{tabular}
\end{table}

%% file: sections/3_method.tex
\section{Method}
\input{figure/pipeline}
\subsection{Problem Formulation}

We study the construction of simulation-ready digital twins from
in-the-wild robot manipulation demonstrations.
Given an RGB video sequence $\mathcal{V}$ and the corresponding robot
trajectory $\mathcal{T}$, WorldMirror constructs a digital twin
\begin{equation}
    \hat{\mathcal{S}} = \mathcal{A}(\mathcal{V}, \mathcal{T}),
\end{equation}
where $\mathcal{A}$ denotes the agentic construction process.
The twin comprises static visual context and support geometry,
editable interactive objects, a calibrated camera, and a robot
configuration assembled in a common coordinate frame.

A simulation-ready digital twin should satisfy three requirements.
First, it should preserve the visual appearance and spatial organization of the environment.
Second, task-relevant objects should be represented as editable
simulation entities with explicit geometry, scale, pose, and
interaction properties.
Third, the scene, objects, and robot should share a consistent,
camera-anchored coordinate frame in which the recorded trajectory
remains executable without motion replanning or retargeting.

WorldMirror formulates this construction process as the iterative
refinement of an executable world program.
During construction, the agent maintains a structured world state
containing reconstructed components, spatial relationships, and
verification feedback.
The router uses this state to select subsequent subtasks and to revise
affected components when verification identifies inconsistencies.

\subsection{Structured World State Memory}

WorldMirror maintains a structured world state as persistent memory
shared by all construction agents.
The state records the current reconstruction, the spatial relationships
among its components, their simulation-relevant attributes, and the
latest verification feedback.
This shared representation allows agents to operate on a consistent
world model and enables the router to condition subsequent decisions
on previously constructed components and detected failures.

Formally, the committed world state at construction step $t$ is
\begin{equation}
    \mathcal{M}_t =
    (\mathcal{S}_t,\mathcal{O}_t,\mathcal{R}_t,\mathcal{F}_t),
\end{equation}
where $\mathcal{S}_t$ denotes the static scene state,
$\mathcal{O}_t$ denotes the interactive object state,
$\mathcal{R}_t$ denotes spatial relationships among scene,
objects, and robot, and $\mathcal{F}_t$ stores verification status
and failure feedback.

Each specialist operates on the current committed state and produces
a candidate update.
The update is committed only when the corresponding verification check
passes.
Otherwise, the committed state remains unchanged, and the typed failure
feedback in $\mathcal{F}_t$ is returned to the router to select the next
specialist based on the diagnosed failure.
Components that depend on a revised component are invalidated and
reconstructed before the updated world is verified again.

\paragraph{Static scene state.}
The static scene state maintains persistent environmental elements,
including the visual background, calibrated camera, and support
geometry required for physical simulation.
We represent scene appearance with a Gaussian representation and model
physical support using corresponding collision geometry.
Together, these components provide the visual and geometric context
for placing interactive objects and executing robot trajectories.

\paragraph{Interactive object state.}
The interactive object state maintains the task-relevant entities
involved in robot manipulation.
Each object is represented by its editable geometry, metric scale,
pose, physical properties, and, when applicable, articulated structure
and interaction attributes.
These properties can be updated independently during construction and
refinement while preserving their geometric and physical relationships
with the static scene and the robot throughout the iterative process.

\subsection{Agentic World Construction}

\paragraph{Router-based Orchestration.}
WorldMirror employs a router agent to supervise the traversal of a
fixed construction graph.
The graph defines the valid ordering and dependencies among construction
stages, including scene preparation, camera calibration, support fitting,
object construction, layout, robot alignment, and trajectory replay.
Rather than freely generating a new pipeline, the router reasons over
the current world state and verification feedback to select among legal
actions, including advancing to the next stage, invoking a construction
skill, rewinding to a predecessor stage, stopping for inspection, or
accepting the reconstructed world.
When verification returns a typed failure with a failure class and a
rewind target, the router follows the corresponding legal transition,
which is validated by the execution controller.
Thus, orchestration is adaptive in its traversal and parameter
selection while preserving the fixed construction dependencies required
for direct trajectory replay.

\paragraph{Agent-as-Tool Delegation.}
WorldMirror adopts an agent-as-tool paradigm, where specialized agents
are exposed as callable capabilities to the router agent.
In our implementation, these capabilities are provided by four
specialized agents: the Reconstruction Agent, Calibration Agent, Layout
Agent, and Replay Agent.
Unlike individual construction skills that perform isolated operations,
the specialized agents handle multi-step subtasks and coordinate the
skills required within their respective domains.
Operating on the shared committed world state, they return candidate
state updates together with task-specific verification results.
The router uses the current state and typed verification feedback to
select the next legal agent transition.

\paragraph{Construction Skills Library.}
WorldMirror maintains a library of reusable construction skills that
support scene reconstruction, asset generation, and camera-anchored
assembly. These skills are invoked by specialized agents to construct
and refine the digital twin. Specifically, \textbf{(a) Camera calibration}
estimates camera parameters that align the reconstructed world with the
real observation. We initialize camera poses using DINOv2-based image
correspondence matching~\citep{oquab2023dinov2} and refine the pose using
a render-to-image objective that combines photometric and structural
alignment, including LSD line-segment matching~\citep{von2012lsd}.
\textbf{(b) Scene and asset reconstruction} generates the visual and physical assets required by the world state. For the static environment, it reconstructs the visual background and corresponding support geometry.
For task-relevant objects, it produces editable simulation assets using methods such as Meshy for rigid objects and Articraft~\citep{zhou2026articraft}
for articulated objects. Finally, \textbf{(c) Camera-anchored assembly} places the reconstructed components in the shared coordinate frame. It refines object scale, position, and orientation to match the observed masks and contours while preserving spatial relationships among the scene,
objects, and robot. The robot base is then aligned to the camera-anchored frame.

\subsection{Verification-driven Refinement}

\paragraph{Iterative Verification.}
WorldMirror verifies each candidate update before committing it to the
world state.
Task-specific verification modules evaluate visual, spatial, and physical
consistency using rendered observations, reconstructed configurations,
and simulation outcomes.
When a discrepancy is detected, the verification module combines these
measurements with the current world state to identify the likely source
of failure.
The verification result is returned to the router as structured feedback,
including the verification status, failure type, rewind target, and
affected components, and is stored in the shared world state to guide
subsequent refinement.

\paragraph{Closed-loop Refinement.}
If the candidate passes its stage-specific check, the update is
committed to the world state.
Otherwise, the previous committed state remains unchanged, and the
router uses the failure class and rewind target to select the next legal
transition in the construction graph.
The affected downstream components are marked outdated and rebuilt after
the failed component has been corrected.
Adaptation therefore occurs at two levels: a specialist searches over
candidate parameters within its assigned stage, while the router
rewinds across dependent stages when the verification evidence requires
a structural correction.
The updated world is then re-evaluated through the same verification
process until all required checks pass or a fixed refinement budget is
reached.

\subsection{Camera Pose-Anchored Coordinate System}

After camera calibration, WorldMirror establishes a unified world frame
$W$ anchored by the calibrated camera.
The static scene, support geometry, interactive objects, and robot base
are all expressed in this frame, so their spatial relationships remain
consistent during assembly and replay.

Let
\[
T^{W}_{C}
=
\begin{bmatrix}
R & t\\
\mathbf{0}^{\top} & 1
\end{bmatrix}
\in SE(3)
\]
denote the transformation from the calibrated camera frame $C$ to the
unified world frame $W$.
A point represented in the camera frame is mapped to the shared world
frame by
\[
X^{W}=T^{W}_{C}X^{C}.
\]
Each reconstructed component is placed in $W$ through its
component-to-world transformation.

The robot base is aligned to the same frame by transforming the recorded camera-to-base extrinsics into $W$, including the required camera-axis conversion and metric scale. Consequently, the robot, scene, support surface, and interactive objects share
a single rigid geometric reference throughout the reconstructed scene. The recorded arm joint trajectory is then executed from the aligned robot base without motion replanning or trajectory retargeting. This camera-anchored formulation preserves the spatial correspondence between the demonstration and the reconstructed simulation, providing the geometric foundation for direct trajectory replay and physical interaction.

%% file: figure/pipeline.tex
\begin{figure}[!t]
    \centering
    \includegraphics[width=\linewidth]{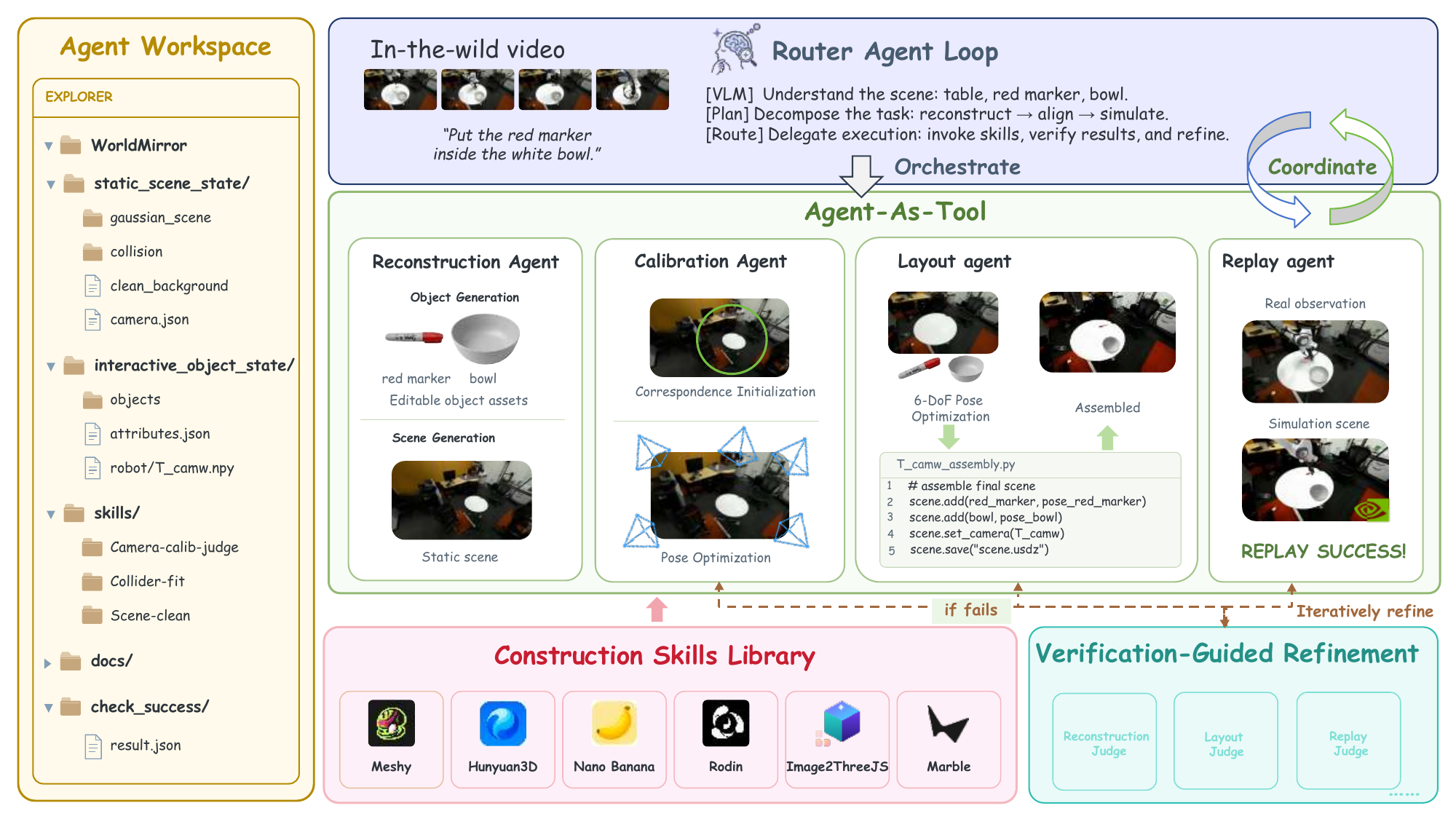}
    \caption{\textbf{WorldMirror framework.} A coding agent maintains an editable world, invokes scene and object generation tools, and aligns camera, assets, and robot through assembly code. Rendered observations and execution feedback guide targeted revisions. The resulting world supports trajectory replay, policy evaluation, and simulation-data generation.}
    \label{fig:framework}
\end{figure}

%% file: sections/4_experiment.tex
\section{Experiments}

We evaluate WorldMirror on in-the-wild robot manipulation demonstrations from DROID~\citep{droid2024}.
Given recorded RGB observations and robot trajectories, WorldMirror
automatically constructs simulation-ready digital twins for direct trajectory replay and interaction through its agentic construction framework. We evaluate the resulting digital twins in terms of visual fidelity, spatial consistency, and interaction consistency against the original demonstrations.

\subsection{Experimental Setup}

\noindent\textbf{Comparisons.}
Our reconstruction baselines are RoLA~\citep{zhao2025rola}, RoboSnap~\citep{zhang2026robosnap}, SimFoundry~\citep{ranawaka2026simfoundry}, and GPT-6 Astra~\citep{openai2026gpt6astra}.
Each table lists the methods evaluated for the corresponding metrics.

\noindent\textbf{Metrics.}
We evaluate digital twin fidelity at three levels: visual fidelity, spatial fidelity, and replay fidelity. 
Visual fidelity is measured using PSNR, SSIM, LPIPS, DISTS, and aesthetic scores; spatial fidelity is evaluated using mask IoU, position error, and relative vector and distance errors; and replay fidelity is measured using object-motion errors, contact F1, and success rate.

\subsection{Mirroring Real-World Appearance}
To evaluate the visual fidelity of the reconstructed digital twins, we compare first-frame simulation renders with real exterior-camera images on ten DROID scenes.
Table~\ref{tab:visual_quality} reports quantitative results, and
Figure~\ref{fig:baselines_comparison} provides comparisons of reconstructed views.
We visualize the wrist-view reconstruction timeline in
Figure~\ref{fig:wrist_timeline}, showing the correspondence between
the observed robot demonstration and the reconstructed simulation
sequence.
WorldMirror achieves the lowest LPIPS and DISTS, together with competitive PSNR and aesthetic scores among evaluated methods.
These results support improved visual correspondence on the evaluated
views, particularly under perceptual metrics.

\input{table/visual_fidelity}
\input{figure/baselines_comparison}
\input{figure/wirst_timeline}

\subsection{Preserving Real-World Spatial Structure}
To assess whether reconstructed scenes preserve real-world spatial
structure, we evaluate object placement and inter-object relationships
on a shared set of 17 task-relevant objects and eight object pairs
from DROID. We report mask IoU, position error, and relative vector and distance
errors, with the 3D metrics computed from visible-surface centroids. As shown in Table~\ref{tab:object_layout_fidelity}, WorldMirror outperforms the evaluated open-source baselines across all four metrics. Compared with GPT-6 Astra, however, it has lower mask IoU and higher position and relative vector errors. WorldMirror achieves the lowest relative distance error, indicating better preservation of pairwise distances on the evaluated scenes, but not a consistent advantage across spatial metrics.

\input{table/Object_layout}

\subsection{Supporting Direct Trajectory Replay and Interaction}
To evaluate whether reconstructed digital twins provide the physical consistency required for real-world manipulation, we replay the original DROID robot trajectories in the reconstructed environments using Isaac Sim without any trajectory modification or additional motion retargeting.
Unlike approaches that require re-planning or inverse kinematics based adaptation to transfer demonstrations into simulation, WorldMirror directly executes the recorded robot trajectories in the reconstructed coordinate system, providing a stricter evaluation of geometric and physical consistency.

We assess motion consistency using displacement, final-position, direction, and DTW errors, and evaluate interaction fidelity using contact F1 and replay success rate.

As shown in Table~\ref{tab:physical_interaction_fidelity}, WorldMirror achieves the lowest errors across all four motion metrics and higher contact F1 than GPT-6 Astra.
It achieves a replay success rate of 100\% on the evaluated demonstrations, compared with 40\% for GPT-6 Astra.
These results demonstrate that the reconstructed digital twins preserve the spatial and physical consistency required for direct trajectory replay and robot-environment interaction without retargeting.

\input{table/physical_interaction}

\subsection{Ablation}
\input{table/ablation-loop}
\input{table/ablation-compoments}

\begin{multicols}{2}

\begin{figure}[H]
    \centering
    \includegraphics[width=0.9\linewidth]{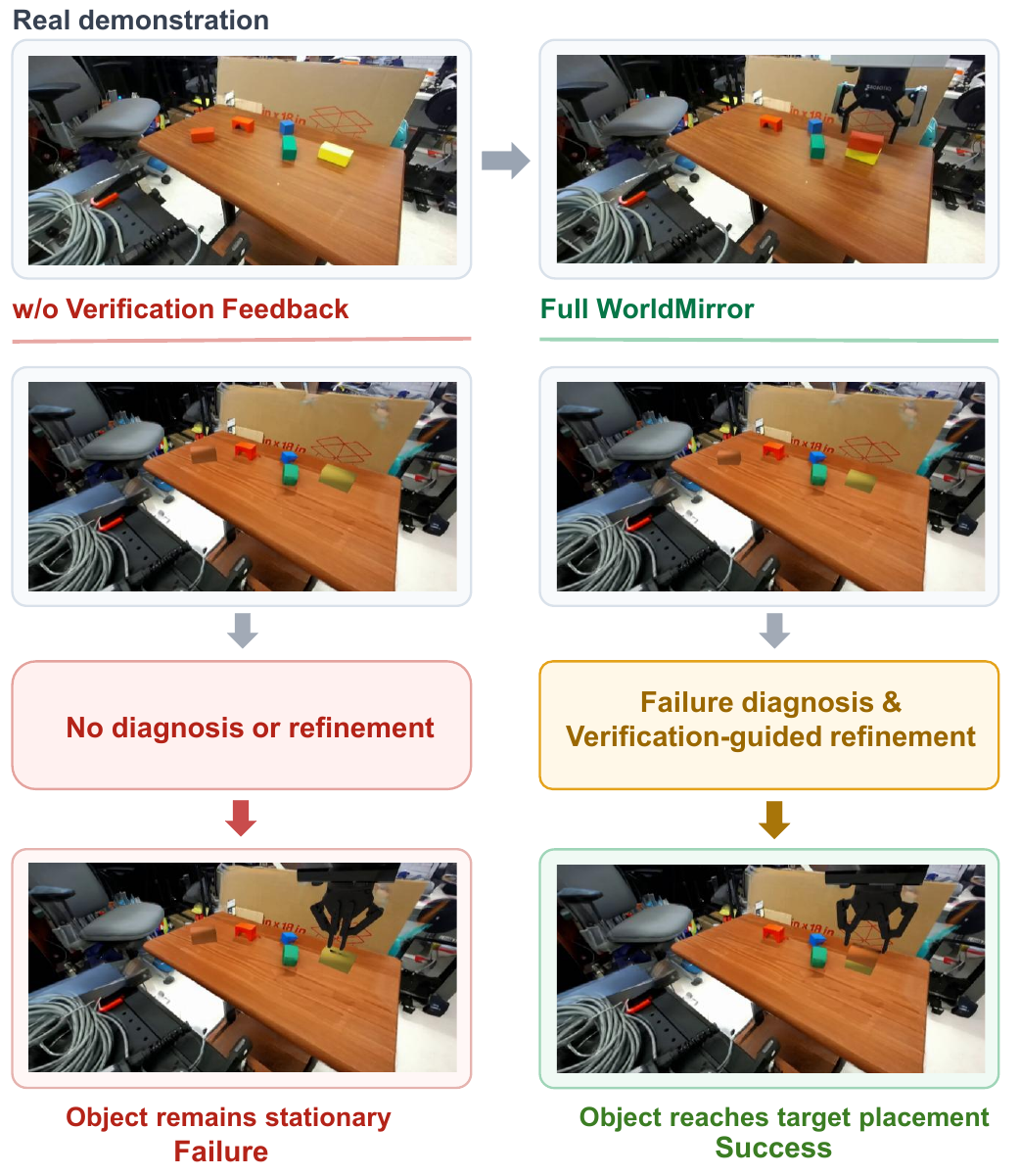}
    \caption{Failure diagnosis and refinement.}
    \label{fig:failure}
\end{figure}

\paragraph{Comparison of refinement strategies.}
To evaluate the effect of different refinement strategies, we compare
three reconstruction strategies in Table~\ref{tab:refinement_ablation}. One-pass performs a single replay, while Fixed Iterative repeatedly
refines the reconstruction without failure-aware diagnosis. In contrast, Full
WorldMirror uses verification feedback to diagnose failures and select targeted
refinement operations. As illustrated in Fig.~\ref{fig:failure}, One-pass fails
to move the object, whereas Full WorldMirror identifies the replay failure and
successfully completes the interaction. Although Fixed Iterative performs
multiple refinement steps, it does not reliably identify the component causing
the failure and therefore fails to recover the interaction. Full WorldMirror
achieves substantially lower replay errors than both non-adaptive variants.
These results show that repeated refinement alone is insufficient, highlighting
the importance of failure-aware diagnosis and targeted refinement.
\columnbreak

\end{multicols}


\paragraph{Contribution of individual components.}
To quantify the contribution of individual system components, we perform component-level ablations on the rail manipulation task under a shared setup with identical replay conditions.
Table~\ref{tab:component_ablation} evaluates stateful memory, adaptive routing, and feedback-driven refinement in terms of replay success and trajectory errors.

Removing stateful memory causes only a minor change, suggesting that its
main benefit lies in preserving intermediate construction results and
verification outcomes across stages. In contrast, removing adaptive routing
leads to replay failure and larger trajectory errors, showing that a fixed
execution sequence cannot reliably respond to intermediate failures.
Removing feedback-driven refinement, equivalent to the One-pass setting,
causes the largest degradation, with the reconstructed object failing to
move successfully. These results identify adaptive routing and verification
feedback as the main components enabling reliable direct trajectory replay,
while stateful memory maintains the context needed for multi-stage
construction.

%% file: table/visual_fidelity.tex
\begin{table}[t]
\centering
\caption{Visual quality comparison on 10 DROID scenes.
Higher is better for $\uparrow$ metrics, while lower is better for $\downarrow$ metrics.}
\label{tab:visual_quality}
\small
\setlength{\tabcolsep}{6pt}
\renewcommand{\arraystretch}{1.2}
\begin{tabular}{@{}llccccc@{}}
\toprule
Type & Method
& PSNR $\uparrow$
& SSIM $\uparrow$
& LPIPS $\downarrow$
& DISTS $\downarrow$
& Aesthetic $\uparrow$ \\
\midrule

\multirow{2}{*}{Open-source}
& RoLA~\citep{zhao2025rola}
& 13.53
& \textbf{0.587}
& 0.506
& 0.206
& 0.480 \\

& RoboSnap~\citep{zhang2026robosnap}
& 10.30
& 0.483
& 0.641
& 0.209
& 0.532 \\

\midrule

Proprietary
& GPT-6 Astra
& 11.90
& 0.515
& 0.606
& 0.299
& 0.472 \\

\midrule

Ours
& \textbf{WorldMirror}
& \textbf{13.57}
& 0.570
& \textbf{0.427}
& \textbf{0.168}
& \textbf{0.570} \\

\bottomrule
\end{tabular}
\end{table}

%% file: figure/baselines_comparison.tex
\begin{figure*}[t]
    \centering
    \includegraphics[width=0.9\textwidth]{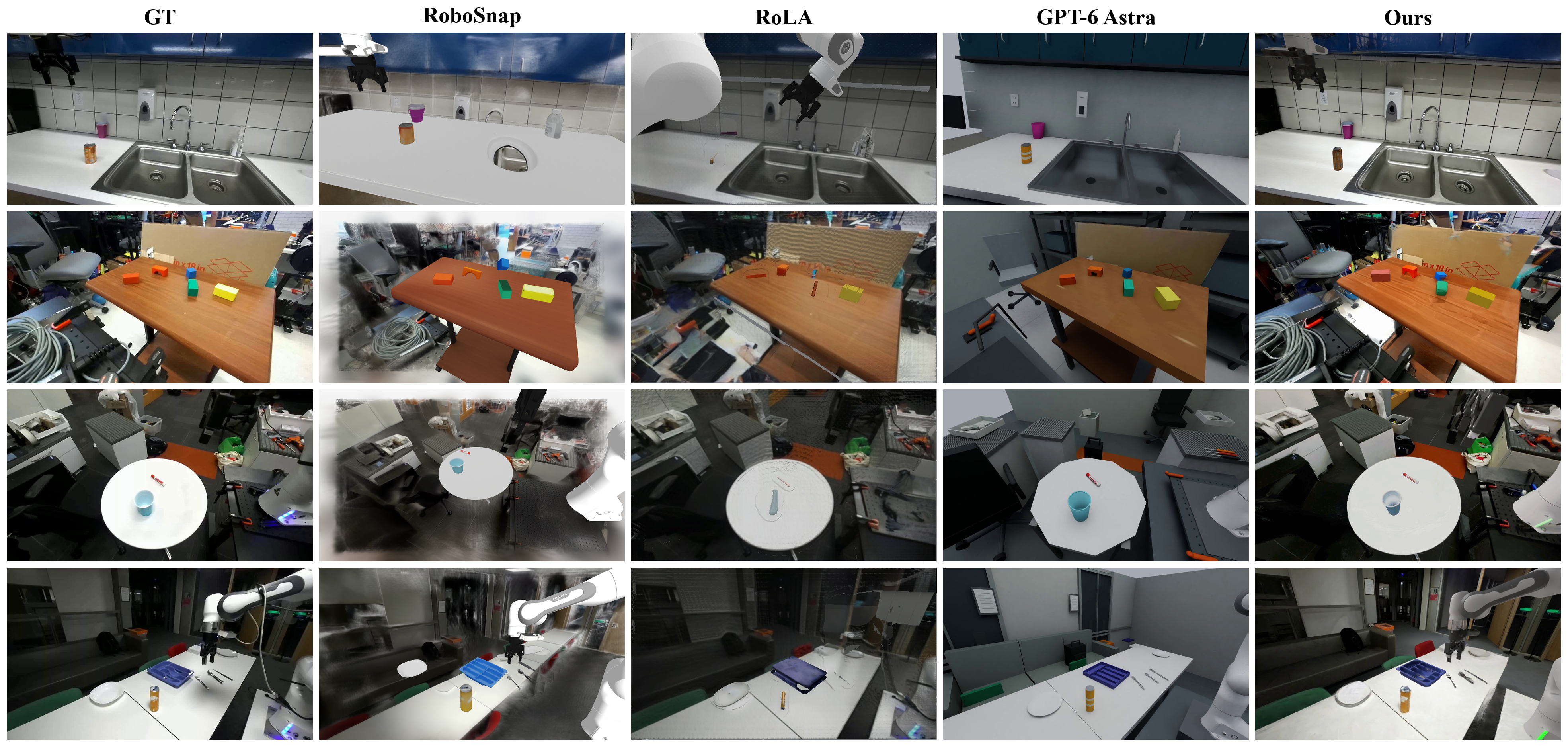}
    \caption{Qualitative RGB comparison with existing baselines on DROID scenes. WorldMirror produces more faithful visual reconstructions with improved appearance consistency and scene details compared with open-source and proprietary baselines.}
    \label{fig:baselines_comparison}
\end{figure*}

%% file: figure/wirst_timeline.tex
\begin{figure*}[t]
    \centering
    \includegraphics[width=0.9\textwidth]{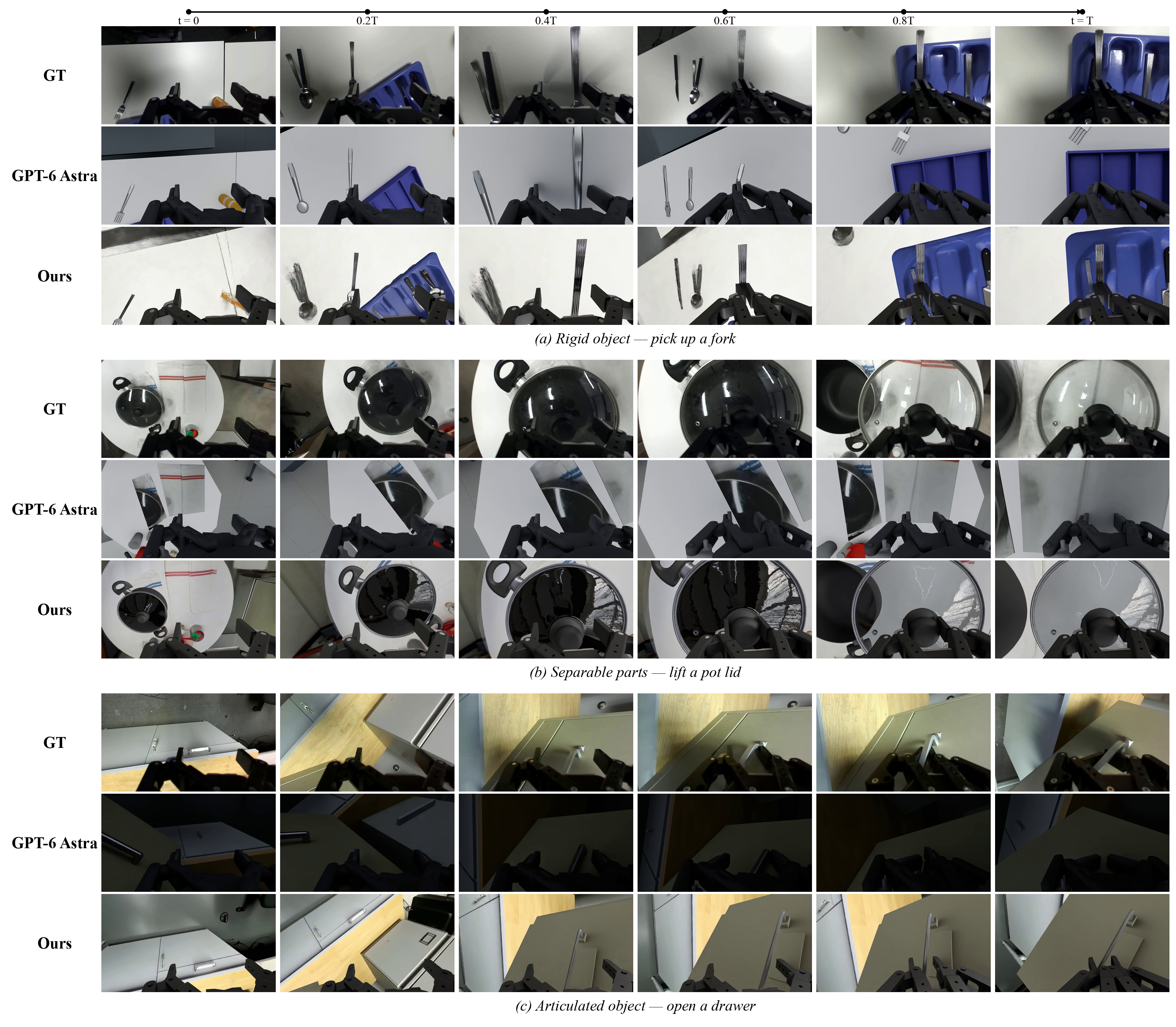}
    \caption{Real-to-sim wrist-view comparison over time, showing visual alignment during replay.}
    \label{fig:wrist_timeline}
\end{figure*}

%% file: table/object_layout.tex
\begin{table}[t]
\centering
\caption{Object layout fidelity comparison. Higher is better for $\uparrow$ metrics, while lower is better for $\downarrow$ metrics.}
\label{tab:object_layout_fidelity}
\small
\setlength{\tabcolsep}{3.5pt}
\renewcommand{\arraystretch}{1.15}
\begin{tabular}{@{}llcccc@{}}
\toprule
Category & Method
& IoU $\uparrow$
& Pos. Err. $\downarrow$
& Rel. Vec. $\downarrow$
& Rel. Dist. $\downarrow$ \\
\midrule

\multirow{3}{*}{Open-source}
& RoLA~\citep{zhao2025rola}
& 0.232
& 30.65
& 0.814
& 0.542 \\

& RoboSnap~\citep{zhang2026robosnap}
& 0.313
& 27.79
& 0.644
& 0.583 \\

& SimFoundry~\citep{ranawaka2026simfoundry}
& 0.169
& 19.54
& 0.627
& 0.507 \\

\midrule

Proprietary
& GPT-6 Astra~\citep{openai2026gpt6astra}
& \textbf{0.598}
& \textbf{8.72}
& \textbf{0.552}
& 0.331 \\

\midrule

Ours
& \textbf{WorldMirror}
& 0.469
& 9.66
& 0.557
& \textbf{0.219} \\

\bottomrule
\end{tabular}
\end{table}

%% file: table/physical_interaction.tex
\begin{table}[t]
\centering
\caption{Motion replay and physical interaction evaluation on DROID demonstrations.
F1 denotes contact F1 score, and SR denotes replay success rate.}
\label{tab:physical_interaction_fidelity}
\footnotesize
\setlength{\tabcolsep}{3.5pt}
\renewcommand{\arraystretch}{1.0}

\begin{tabular}{@{}lcccccc@{}}
\toprule
& \multicolumn{4}{c}{Motion Replay}
& \multicolumn{2}{c}{Physical Interaction} \\
\cmidrule(lr){2-5}
\cmidrule(lr){6-7}
Method
& Disp. (cm)$\downarrow$
& Final (cm)$\downarrow$
& Dir. ($^\circ$)$\downarrow$
& DTW (cm)$\downarrow$
& F1$\uparrow$
& SR$\uparrow$ \\
\midrule

RoLA~\citep{zhao2025rola}
& 24.70
& 40.80
& 53.00
& 10.30
& --
& -- \\

GPT-6 Astra~\citep{openai2026gpt6astra}
& 11.17
& 21.98
& 81.30
& 10.82
& 0.653
& 40\% \\

\textbf{WorldMirror}
& \textbf{1.79}
& \textbf{5.83}
& \textbf{10.90}
& \textbf{1.64}
& \textbf{0.873}
& \textbf{100\%} \\

\bottomrule
\end{tabular}
\end{table}

%% file: table/ablation-loop.tex
\begin{table}[t]
\centering
\caption{
Effect of verification-driven adaptive refinement on replay performance.
One-pass performs reconstruction followed by a single replay attempt.
Fixed Iterative repeatedly applies refinement without failure-aware diagnosis.
Full WorldMirror performs structured failure diagnosis and targeted refinement.
}
\label{tab:refinement_ablation}

\small
\setlength{\tabcolsep}{4pt}
\renewcommand{\arraystretch}{1.2}

\begin{tabular}{@{}lccccccc@{}}
\toprule
Strategy
& Success $\uparrow$
& Final (cm)$\downarrow$
& Trans. (cm)$\downarrow$
& Lift (cm)$\downarrow$
& Dir. ($^\circ$)$\downarrow$
& DTW (cm)$\downarrow$
& Replay Iter. $\downarrow$
\\
\midrule

One-pass
& \xmark
& 33.86
& 0
& 0
& --
& 14.88
& 1
\\

Fixed Iterative
& \xmark
& 31.77
& 2.16 
& 0.40
& 13.9
& 13.90
& 2
\\

\textbf{Full}
& \cmark
& \textbf{2.34}
& \textbf{32.98}
& \textbf{11.11}
& \textbf{3.70}
& \textbf{0.95}
& 4
\\

\bottomrule
\end{tabular}
\end{table}

%% file: table/ablation-compoments.tex
\begin{table*}[t]
\centering
\caption{
Component ablation of WorldMirror on the rail manipulation task.
The ablations remove stateful memory, adaptive routing, or
feedback-driven refinement. The latter corresponds to the One-pass
setting, which performs only one replay without subsequent refinement.
}
\label{tab:component_ablation}

\small
\setlength{\tabcolsep}{4pt}
\renewcommand{\arraystretch}{1.2}

\resizebox{0.98\textwidth}{!}{%
\begin{tabular}{@{}lcccccc@{}}
\toprule
\textbf{Variant}
& \textbf{Success} $\uparrow$
& \textbf{Final} (cm) $\downarrow$
& \textbf{Trans.} (cm) $\downarrow$
& \textbf{Lift} (cm) $\downarrow$
& \textbf{Dir.} ($^\circ$) $\downarrow$
& \textbf{DTW} (cm) $\downarrow$
\\
\midrule

\textbf{Full}
& \textbf{\cmark}
& \textbf{2.34}
& \textbf{32.98}
& \textbf{11.11}
& \textbf{3.70}
& \textbf{0.95}
\\

w/o Stateful Memory
& \cmark
& 2.35
& 33.00
& 11.23
& 3.75
& 0.97
\\

w/o Adaptive Routing
& \xmark
& 9.86
& 42.23
& 6.45
& 7.93
& 2.60
\\

w/o Feedback Refinement
& \xmark
& 33.86
& 0
& 0
& --
& 14.88
\\

\bottomrule
\end{tabular}%
}

\end{table*}

%% file: sections/5_conclusion.tex
\section{Conclusion}

We introduced \textbf{WorldMirror}, an agentic real-to-sim framework that
constructs simulation-ready digital twins from in-the-wild robot demonstrations.
By combining structured world state, specialized construction agents, and
verification-driven refinement, WorldMirror preserves the spatial and physical
consistency needed to replay recorded trajectories without modification.
Experiments on DROID demonstrate improved spatial alignment, motion replay, and
interaction fidelity.
These results highlight the importance of coupling visual reconstruction with
physical verification and trajectory-level evaluation.
They also show that executable world states can provide a useful interface
between real demonstrations and simulation-based robot learning.

Our evaluation remains limited to a small set of DROID demonstrations and does
not yet include downstream policy learning, real-world deployment, or broader
cross-task generalization.
The reconstruction quality also depends on the available reconstruction and
simulation tools.
Future work will study more diverse tasks and the use of reconstructed twins
for downstream robot learning.